\documentclass[conference]{IEEEtran}
\IEEEoverridecommandlockouts
\usepackage{cite}
\usepackage{amsmath, epsfig, ulem, amssymb, bm, bbm}
\usepackage{hyperref}
\usepackage{subfigure}
\usepackage{booktabs} 
\usepackage{bbding}
\usepackage{graphicx}
\usepackage{float}
\usepackage{array}
\usepackage{caption}
\usepackage{multirow}
\usepackage{enumitem}
\usepackage{multicol}
\usepackage{tabularx}
\usepackage{threeparttable}
\usepackage{verbatim}
\usepackage{bbding}

\usepackage{xcolor}
\usepackage{stfloats}
\usepackage[linesnumbered,ruled,vlined]{algorithm2e}
\SetKwInput{KwInput}{Input}                
\SetKwInput{KwOutput}{Output}              

\usepackage{amsmath,amssymb,amsfonts}
\usepackage{algorithmic}
\usepackage{graphicx}
\usepackage{textcomp}
\usepackage{xcolor}
\def\BibTeX{{\rm B\kern-.05em{\sc i\kern-.025em b}\kern-.08em
    T\kern-.1667em\lower.7ex\hbox{E}\kern-.125emX}}
\begin{document}

\title{Deep Metric Multi-View Hashing for Multimedia Retrieval}

\author{\IEEEauthorblockN{Jian Zhu}
\IEEEauthorblockA{Zhejiang Lab \\
Hangzhou, Zhejiang, China \\
qijian.zhu@zhejianglab.com
}\\   
\IEEEauthorblockN{Zhangmin Huang}
\IEEEauthorblockA{Zhejiang Lab \\
Hangzhou, Zhejiang, China \\
zmhuang@zhejianglab.com
}\\   
\and
\IEEEauthorblockN{Xiaohu Ruan}
\IEEEauthorblockA{vivo AI lab \\
Hangzhou, Zhejiang, China \\
xiaohu.ruan@vivo.com
}
 \\ 
\IEEEauthorblockN{Yu Cui}
\IEEEauthorblockA{Zhejiang Lab \\
Hangzhou, Zhejiang, China \\
cui.yu@zhejianglab.com
}
\and
\IEEEauthorblockN{Yongli Cheng}
\IEEEauthorblockA{Fuzhou University \\
Fuzhou, Fujian, China \\
chengyongli@fzu.edu.cn
} \\ 
\IEEEauthorblockN{%
Lingfang Zeng\IEEEauthorrefmark{1}\thanks{\IEEEauthorrefmark{1}Lingfang Zeng is the corresponding author.}}
\IEEEauthorblockA{Zhejiang Lab \\
Hangzhou, Zhejiang, China \\
zenglf@zhejianglab.com
}

}

\maketitle

\begin{abstract}
Learning the hash representation of multi-view heterogeneous data is an important task in multimedia retrieval. However, existing methods fail to effectively fuse the multi-view features and utilize the metric information provided by the dissimilar samples, leading to limited retrieval precision. Current methods utilize weighted sum or concatenation to fuse the multi-view features. We argue that these fusion methods cannot capture the interaction among different views. Furthermore, these methods ignored the information provided by the dissimilar samples. We propose a novel deep metric multi-view hashing (DMMVH) method to address the mentioned problems. Extensive empirical evidence is presented to show that gate-based fusion is better than typical methods. We introduce deep metric learning to the multi-view hashing problems, which can utilize metric information of dissimilar samples. On the MIR-Flickr25K, MS COCO, and NUS-WIDE, our method outperforms the current state-of-the-art methods by a large margin (up to $15.28$ mean Average Precision (mAP) improvement).
\end{abstract}

\begin{IEEEkeywords}
Multi-view hash, Multi-modal hash, Deep metric learning, Multimedia retrieval
\end{IEEEkeywords}

\section{Introduction}
Multi-view hashing is utilized to solve multimedia retrieval problems. A well-designed multi-view hashing algorithm can dramatically improve the precision of multimedia retrieval tasks. Different from single-view hashing, which only searches in a single-view way, multi-view hashing can utilize data from different sources (e.g., image, text, audio, and video). Multi-view hashing representation learning first extracts heterogeneous features from different views, then fuses multi-view features to capture a global representation of different views. 
\begin{figure}
	\centering
	\includegraphics[width=9cm]{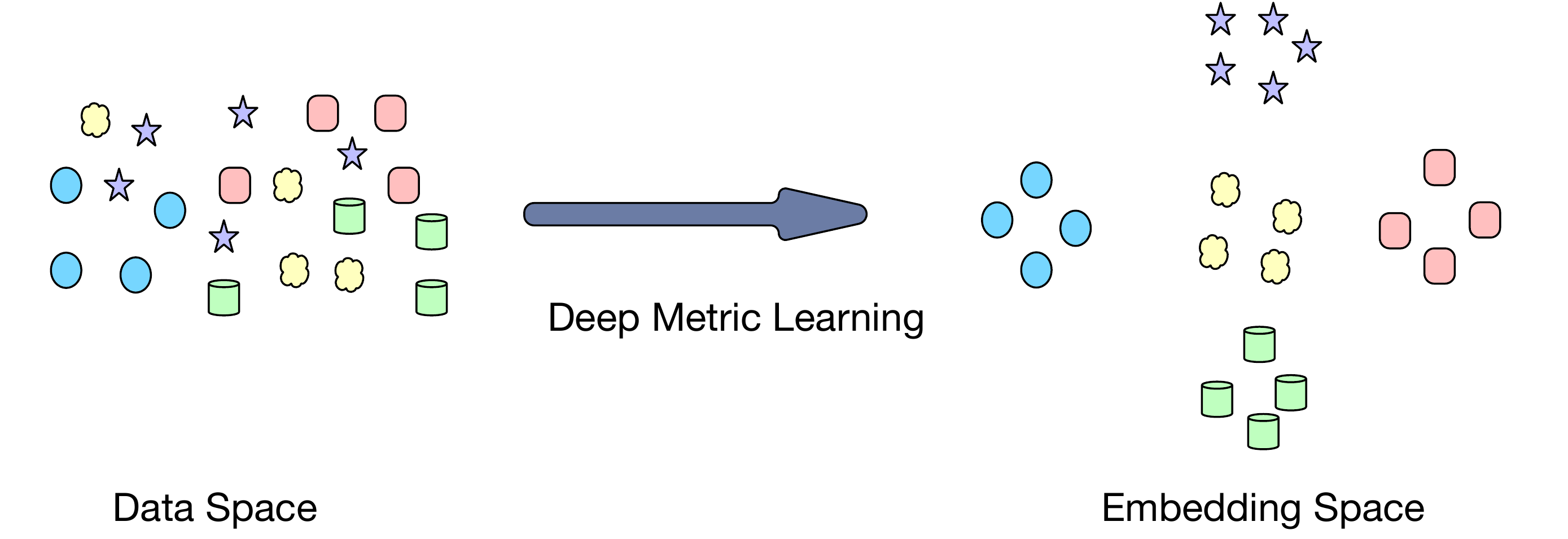}
	\caption{A schematic of deep metric learning. The inputs are randomly distributed in the data space. Deep metric learning projects the inputs to the embedding space, where the embeddings are allocated concerning their semantic meaning.}
	\label{fig:01}
\end{figure}

Current multi-view hashing algorithms suffer from low retrieval precision. It is mainly caused by the following two aspects. First, the fusion of multi-view features is insufficient for current multi-view hashing algorithms. To get a global representation, typical multi-view hashing methods (e.g., Deep Collaborative Multi-View Hashing) (DCMVH) \cite{zhu:17}, Flexible Multi-modal Hashing (FMH) \cite{zhu:52}) utilizes weighted sum or concatenation to fuse the multi-view features. The relationship between the texts and images is ignored during the fusing process, which incurs a weak expressiveness of the obtained global representation. Second, current methods are confined by the information provided by similar samples. The importance of measuring the distance between dissimilar samples is underrated. For instance, Flexible Graph Convolutional Multi-modal Hashing (FGCMH) \cite{lu:18} is a GCN-based \cite{welling:50} multi-view hashing method, which constructs the edges of a graph based on similarity and aggregates features of adjacent nodes. Hence, dissimilar samples do not play a role during this procedure. 

We propose a \textit{Deep Metric Multi-View Hashing} method termed DMMVH. It takes advantage of Context Gating \cite{miech:54} to learn the interaction and dependency between the image and text features. Unlike typical methods, DMMVH fuses multi-view features into a global representation without losing dependency on these features. Moreover, deep metric learning is introduced to DMMVH. As shown in Fig. \ref{fig:01}, initially, samples are distributed randomly in the raw data space. Using deep metric learning, semantically similar samples are close to one another, while dissimilar samples are pushed away. To utilize the distance information of dissimilar samples, we design a deep metric loss function. Furthermore, we introduce a hyper-parameter to reduce the complexity of the designed loss function. The optimal embedding space is obtained through deep metric learning, which follows the semantics-preservation principle of hash representation learning. 

We evaluate our method on MIR-Flickr25K, MS COCO, and NUS-WIDE datasets in multi-view hash representation learning benchmarks. The proposed method provides up to $15.28\%$ mAP improvement in benchmarks.

Our main contributions are as follows:

\begin{itemize}
\item We propose a novel multi-view hash method, which achieves state-of-the-art results in multimedia retrieval.
\item We take advantage of Context Gating to learn a better global representation of different views to address the insufficient fusion problem.
\item Deep metric learning is introduced to multi-view hashing for the first time. A deep metric loss with linear complexity is designed and optimized.

\end{itemize}

\begin{figure*}
	\centering
	\includegraphics[width=18cm]{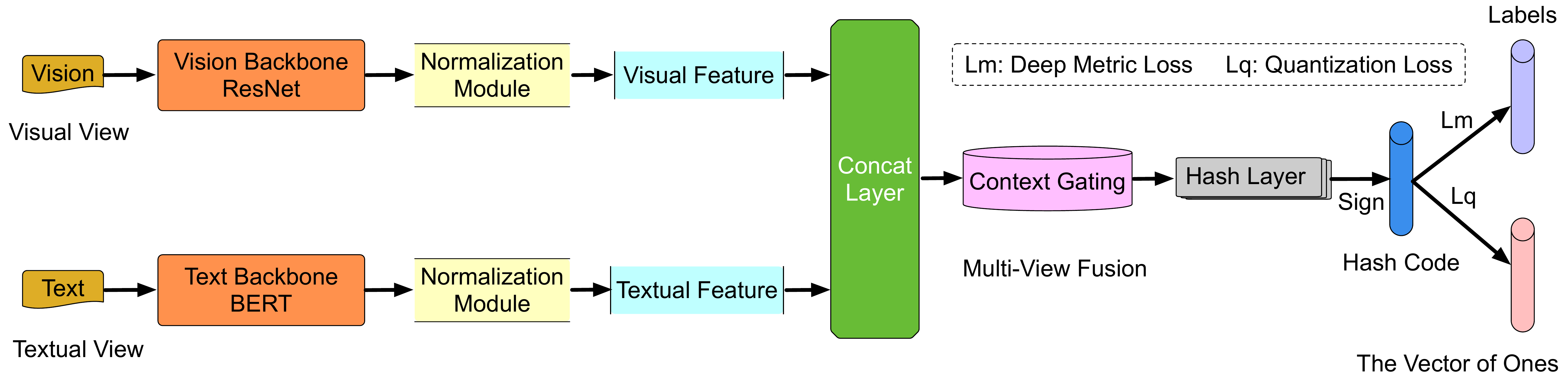}
	\caption{The flow chart of the DMMVH method. The image and text features are extracted by ResNet and BERT, respectively. The features are normalized by the normalization module and concatenated together. Multi-view fusion module performs Context Gating on the concatenated features and fuses multi-view features while preserving the dependency. Finally, the hash layer produces a hash code based on the fused representation.}
	\label{fig:02}
\end{figure*}

\section{The Proposed Methodology}
DMMVH aims to utilize a newly designed deep metric loss to train a deep multi-view hashing network. We first present the deep multi-view hashing network, which deeply fuses the multi-view features into a global representation. Then the new deep metric loss is turned to illustrate. Eventually, a hyper-parameter $\lambda$ is introduced to reduce the complexity. 
\subsection{Deep Multi-View Hashing Network}
\label{section:proposed_method} 
Deep multi-view hashing network is designed to convert multi-view data into hash code. As shown in Fig. \ref{fig:02}, DMMVH consists of a vision backbone, text backbone, normalization modules, multi-view fusion module, and a hash layer. These modules are described in detail below.
\begin{enumerate}
\item \textbf{Vision Backbone:} Deep ResNet \cite{he:5} is employed to produce visual features. 

\item \textbf{Text Backbone:} The BERT-base \cite{devlin:3} is utilized to extract text features. 

\item \textbf{Normalization Module:} Normalization module projects multi-view features (visual and text features) into the same dimension and threshold. 

\item \textbf{Multi-View Fusion Module:}
We employ Context Gating to fuse the concatenated visual and text features. The multi-view fusion module projects the input multi-view features into a new global representation as:
\begin{equation}
	X_{\text{fusion}} =\sigma(w_{\text{fusion}}X_{\text{concat}}+b_{\text{fusion}})\circ X_{\text{concat}},
\end{equation}
where $X_{\text{concat}} \in \mathbb{R}^{n}$ is the multi-view feature vector, $\sigma$ is the element-wise sigmoid activation, and $\circ$  is the element-wise multiplication.  $w_{\text{fusion}} \in \mathbb{R}^{n \times n}$  and $b_{\text{fusion}} \in \mathbb{R}^{n}$ are trainable parameters. The vector of weights $\sigma(w_{\text{fusion}}X_{\text{concat}}+b_{\text{fusion}}) \in [0, 1]$ represents a set of learned gates applied to the individual dimensions of the input feature $X_{\text{concat}}$.

\item \textbf{Hash Layer:} A linear layer with a $\tanh$ activation is hired as the hash layer, which can be represented as $h_{\text{k-bit}} = \text{sgn}[\tanh(w_{\text{hash}}X_{\text{fusion}}+b_{\text{hash}})]$,
where $sgn$ represents the signum function. $w_{\text{hash}} \in \mathbb{R}^{n \times n}$  and $b_{\text{hash}} \in \mathbb{R}^{n}$ are trainable parameters. The output has the same number of dimensions as the hash code. 
\end{enumerate}
\subsection{Deep Metric Loss}
Assume that the training dataset $\mathcal{X}= \left\{\left\{(x_{i}, y_i) \right\}_{i=1}^{N}\right\}$, where $x_{i} \in \mathbb{R}^{D}$ is a multi-view instance and $y_{i}$ denotes the category information of $x_{i}$. Furthermore, $F: x \mapsto h$ denotes the deep multi-view hashing network, which maps the input space $\mathbb{R}^{D}$ to K-bit Hamming space $\{-1,1\}^{K}$. 

Let $h_{i} = F(x_i) $ be the hash code of $x_i$. Then, we have an elegant linear relationship between Hamming distance ${dist}_{H}(\cdot, \cdot)$ and inner product $\langle\cdot, \cdot\rangle$
\begin{equation}\label{hamming}
{dist}_{H}\left(h_{i}, h_{j}\right)=\frac{1}{2}\left(K-\phi_{ij}\right),
\end{equation}
where $\phi_{ij} = \left\langle h_{i}, h_{j}\right\rangle$. For $x_i$, its label $y_{i} \in\{0,1\}^{C}$ and ${C}$ is the number of categories. Notice that, one sample may belong to multiple categories. Given the semantic label information, the pairwise similarity matrix $S=\left\{s_{i j}\right\}$ can defined as follows: if $x_{i}$ and $x_{j}$ are semantically similar then $s_{i j}=1$, otherwise, $s_{i j}=0$.

Provided the matrix $\Phi = (\phi_{ij})$ and $S = (s_{ij})$,
combining the cross-entropy loss and deep metric learning yields the loss function
\begin{equation}
	L'_{m}=\frac{1}{N^2}\sum\limits_{i=1}^{N}\sum\limits_{j=1}^{N}[s_{ij} \log (1+e^{{\phi}_{ij} })-s_{ij} \phi_{ij}].
	\label{eq:Lm0}
\end{equation}

Since $s_{ij}$  can only be $0$ or $1$, when $s_{ij} = 0$, the loss $L'_{m}$ vanishes, which means the dissimilar samples do not play any role in the training. Notice that, the first part of the metric loss is $\log(1+e^{{\phi}_{ij}})$. Considering the elegant linear relationship between Hamming distance and the inner product, i.e., Eq. \eqref{hamming}, as the inner product $\phi_{ij}$ decreases, the Hamming distance will increases. Therefore, this part is a proper metric loss. It punishes the dissimilar samples having a closer distance in the embedding space while rewarding a larger distance between them. Due to the above analysis, we revise Eq. \eqref{eq:Lm0} as
\begin{equation}\label{eq:Lm1}
	L_{m}=\frac{1}{N^2}\sum\limits_{i=1}^{N}\sum\limits_{j=1}^{N}[w_{d}\log (1+e^{{\phi}_{ij} })-s_{ij} \phi_{ij}].
\end{equation}
$w_{d}$ represents the loss weight of dissimilar sample pairs. With this revising, the dissimilar samples can also help the training. The derivation of the metric loss can be found in the appendix.

\subsection{Hyper-parameter $\lambda$}
Notice that, calculating the matrix $\Phi$ or $S$ is $O(N^2)$ complexity. By introducing a hyper-parameter $\lambda$, calculating any one of them only has $O(\lambda^2bN)$ complexity, where $b$ is the batch size. 

We randomly choose a portion of them to calculate the similarity matrix, instead of calculating a global similarity matrix $S$ for every sample. Assume the samples are already shuffled. Let $b$ be the batch size and $\lambda$ be a hyper-parameter. We take the first $\lambda b$ and last $\lambda b$ samples to calculate the loss. Specifically, let $H_{\text{prec}} =\{h_{1}, h_{2}, \ldots, h_{\lambda b}\}$, 
$H_{\text{rest}} =\{h_{(1-\lambda)b+1}, h_{(1-\lambda)b+2}, \ldots, h_{b}\}$,
$Y_{\text{prec}} =\{y_{1}, y_{2}, \ldots, y_{\lambda b}\}$, 
and $Y_{\text{rest}} =\{y_{(1-\lambda) b+1}, y_{(1-\lambda) b+2}, \ldots, y_{b}\}$. Then we have two matrices: $\Phi_{batch}$  and $S_{batch}$

\begin{equation}
	\Phi_{\text{batch}} =H_{\text{prec}} \times  H_{\text{rest}}^{T} =\begin{bmatrix} \phi_{ij}  \end{bmatrix}
\end{equation}

\begin{equation}
	S_{\text{batch}} =Y_{\text{prec}} \times  Y_{\text{rest}}^{T}  =\begin{bmatrix} s_{ij}   \end{bmatrix},
\end{equation}
where $\times$  represents the matrix multiplication operation. With this designing, Eq. \eqref{eq:Lm1} reduces to
\begin{equation}
	L_{m}=\frac{1}{(\lambda b)^2}\sum\limits_{i=1}^{\lambda b}\sum\limits_{j=(1-\lambda)b+1}^{b}[w_{d}\log (1+e^{{\phi}_{ij} })-s_{ij} \phi_{ij}].
	\label{eq:Lm2}
\end{equation}

Eventually, a quantization loss is introduced to refine the generated hash codes, which can be represented as:
\begin{equation}
	L_{q}=\frac{1}{b}\sum_{i\in I}^{} ({\||\boldsymbol{h}_{i}|-\mathbf{1}\|_{2}}),
\end{equation}
where $I= \{i \mid 1\le i \leq \lambda b, i \in \mathbb{N}\}\cup \{i \mid(1-\lambda) b+1\le i \leq b, i \in \mathbb{N}\}$. Combining the metric loss and the quantization loss by weighted sum yields the total loss function of our method
\begin{equation}
	L_{\text{Total}}= L_{m}+\mu L_{q},
	\label{eq:loss}
\end{equation}
where $\mu$ is a hyper-parameter obtained through grid search in our work. 

\section{Experiments}
Extensive experiments are conducted to evaluate the proposed DMMVH method against eleven state-of-the-art multi-view hashing methods on three public benchmark datasets.

\textbf{Datasets:} Three genetic datasets are adopted: MIR-Flickr25K \cite{huiskes:20}, NUS-WIDE \cite{chua:22}, and MS COCO \cite{lin:21}. These datasets have been widely used for evaluating multimedia retrieval performance. The statistics of three datasets are summarized in Table \ref{Tab:01}.

\textbf{Evaluation Metric:} We utilize the mean Average Precision (mAP) as the evaluation metric. 
\begin{table*}
	\centering
   \caption{General statistics of three datasets. The dataset size, number of categories, and feature dimensions are included.}
	\begin{tabular}{llllllll}
		\toprule[1pt]
		Datasets   & Training Size & Retrieval Size & Query Size & Categories&Visual Feature & Textual Feature \\ \midrule[0.8pt]
		MIR-Flickr25K & 5000  & 17772 & 2243    & 24&ResNet(768-D) &BERT(768-D)\\
		MS COCO & 18000  & 82783 & 5981    & 80&ResNet(768-D) &BERT(768-D)\\
		NUS-WIDE & 21000  & 193749 & 2085    & 21&ResNet(768-D) &BERT(768-D)\\
		\bottomrule[1pt]
	\end{tabular}

	\label{Tab:01}
\end{table*}

\textbf{Baseline:} To evaluate the retrieval performance, the proposed method is compared with eleven multi-view hashing methods, including four unsupervised methods (MFH \cite{song:9}, MAH  \cite{liu:10}, MVLH \cite{shen:6}, and MvDH \cite{shen:11})  and seven supervised methods (MFKH \cite{liu:7}, DMVH  \cite{yang:12}, FDMH \cite{liu:23}, FOMH \cite{lu:24}, DCMVH \cite{zhu:17},  SAPMH \cite{zheng:25}, and FGCMH \cite{lu:18}). 

\textbf{Implementation Details:}
Our implementation is on the PyTorch platform. For the feature extraction backbones, we use the pre-trained models, specifically ResNet-50 and BERT-base. The dropout probability is set to be $0.1$ to improve the generalization capability. We employ the AdamW optimizer with an initial learning rate $1 \times 10^{-5}$ and set $\beta_{1} = 0.9$, $\beta_{2} = 0.999$. The hyper-parameter $\lambda$ of the loss function for deep metric learning is $0.5$. The combination coefficient $\mu$ of the total loss function is set to be $0.5$. Let the loss weight $w_{d}$ of dissimilar sample pairs be $1.5$.

\subsection{Analysis of Experimental Results}
\begin{table*}
	\setlength{\tabcolsep}{2pt}
	\centering
	\caption{mAP Comparison Results on MIR-Flickr25K, NUS-WIDE, and MS COCO. The best results are bolded, and the previous state-of-the-art results are underlined.  The * indicates that the results on this dataset are of statistical significance.}
	\begin{tabular}{llllllllllllll}
		\toprule[1pt]
		\multicolumn{1}{c}{\multirow{2}{*}{Methods}} & \multicolumn{1}{c}{\multirow{2}{*}{Ref.}} & \multicolumn{4}{c}{MIR-Flickr25K*}    & \multicolumn{4}{c}{NUS-WIDE*}      & \multicolumn{4}{c}{MS   COCO*}       \\  \cmidrule(r){3-6}  \cmidrule(r){7-10}  \cmidrule(r){11-14}
		\multicolumn{1}{c}{}                         & \multicolumn{1}{c}{}                      & 16 bits & 32 bits & 64 bits & 128 bits & 16 bits & 32 bits & 64 bits & 128 bits & 16 bits & 32 bits & 64 bits & 128 bits \\ \midrule[0.8pt]
		MFH                                          & TMM13                                     & 0.5795 & 0.5824 & 0.5831 & 0.5836  & 0.3603 & 0.3611 & 0.3625 & 0.3629  & 0.3948 & 0.3699 & 0.3960  & 0.3980   \\
		MAH                                          & TIP15                                     & 0.6488 & 0.6649 & 0.6990  & 0.7114  & 0.4633 & 0.4945 & 0.5381 & 0.5476  & 0.3967 & 0.3943 & 0.3966 & 0.3988  \\
		MVLH                                         & MM15                                      & 0.6541 & 0.6421 & 0.6044 & 0.5982  & 0.4182 & 0.4092 & 0.3789 & 0.3897  & 0.3993 & 0.4012 & 0.4065 & 0.4099  \\
		MvDH                                         & TIST18                                    & 0.6828 & 0.7210  & 0.7344 & 0.7527  & 0.4947 & 0.5661 & 0.5789 & 0.6122  & 0.3978 & 0.3966 & 0.3977 & 0.3998  \\ \midrule[0.8pt]
		MFKH                                         & MM12                                      & 0.6369 & 0.6128 & 0.5985 & 0.5807  & 0.4768 & 0.4359 & 0.4342 & 0.3956  & 0.4216 & 0.4211 & 0.4230  & 0.4229  \\
		DMVH                                         & ICMR17                                    & 0.7231 & 0.7326 & 0.7495 & 0.7641  & 0.5676 & 0.5883 & 0.6902 & 0.6279  & 0.4123 & 0.4288 & 0.4355 & 0.4563  \\
		FOMH                                         & MM19                                      & 0.7557 & 0.7632 & 0.7564 & 0.7705  & 0.6329 & 0.6456 & 0.6678 & 0.6791  & 0.5008 & 0.5148 & 0.5172 & 0.5294  \\
		FDMH                                         & NPL20                                     & 0.7802 & 0.7963 & 0.8094 & 0.8181  & 0.6575 & 0.6665 & 0.6712 & 0.6823  & 0.5404 & 0.5485 & 0.5600   & 0.5674  \\
		DCMVH                                        & TIP20                                     & 0.8097 & 0.8279 & 0.8354 & 0.8467  & 0.6509 & 0.6625 & 0.6905 & \underline{0.7023}  & 0.5387 & 0.5427 & 0.5490  & 0.5576  \\
		SAPMH                                        & TMM21                                      & 0.7657 & 0.8098 & 0.8188 & 0.8191  & 0.6503 & 0.6703 & 0.6898 & 0.6901  & 0.5467 & \underline{0.5502} & 0.5563 & 0.5672  \\
		FGCMH                                        & MM21                                      & \underline{0.8173} & \underline{0.8358} & \underline{0.8377} & \underline{0.8606}  & \underline{0.6677} & \underline{0.6874} & \underline{0.6936} & 0.7011  & \underline{0.5641} & 0.5273 & \underline{0.5797} & \underline{0.5862}  \\\midrule[0.8pt]
		DMMVH                                         & Proposed                                         &\textbf{0.8587} & \textbf{0.8707} & \textbf{0.8798} & \textbf{0.8827} & \textbf{0.7714} & \textbf{0.7820} & \textbf{0.7879} & \textbf{0.7916}  &\textbf{0.6716} & \textbf{0.7030} & \textbf{0.7122} & \textbf{0.7244} \\
		\bottomrule[1pt]
	\end{tabular}

	\label{Tab:02}
\end{table*}

\begin{table*}[!t]
	\setlength{\tabcolsep}{2pt}
	\centering
	\caption{Ablation Experiments On Three Datasets. Effects of Deep Multi-View Hash Network Architecture.}
	\begin{tabular}{lllllllllllll}
		\toprule[1pt]
		\multicolumn{1}{c}{\multirow{2}{*}{Methods}} & \multicolumn{4}{c}{MIR-Flickr25K}   & \multicolumn{4}{c}{NUS-WIDE}  & \multicolumn{4}{c}{MS COCO} \\   \cmidrule(r){2-5}  \cmidrule(r){6-9}  \cmidrule(r){10-13}
		\multicolumn{1}{c}{} & 16 bits & 32 bits & 64 bits & 128 bits & 16 bits & 32 bits & 64 bits & 128 bits& 16 bits & 32 bits & 64 bits & 128 bits \\  \midrule[0.8pt] 
		DMMVH-metric    & 0.8531 &  0.8614 &  0.8708 &  0.8738  & 0.7671 & 0.7766 & 0.7809 & 0.7872& 0.6686& 0.6970&0.7066&0.7078\\
		DMMVH-quant    & 0.5531 &  0.5531 &  0.5531 &  0.5531  & 0.3085 & 0.3085 & 0.3085 & 0.3085 &0.3502&0.3502&0.3502&0.3502\\
		DMMVH-text    & 0.6047 &  0.6107 &  0.6104 &  0.6119  & 0.3623 & 0.3585 & 0.3649 & 0.3631 &0.5819&0.5886&0.5955&0.5992\\
		DMMVH-image    & 0.8292 &  0.8425 &  0.8547 &  0.8631  & 0.7530 & 0.7593 & 0.7689 & 0.7778 & 0.6598&0.6886&0.7033&0.7160    \\
		DMMVH-concat    & 0.8498 &  0.8633 &  0.8742 &  0.8777  & 0.7635 & 0.7713 & 0.7827 & 0.7866   & 0.6615&0.6932&0.7056&0.7188 \\ \midrule[0.8pt]
		DMMVH    &\textbf{0.8587} & \textbf{0.8707} &\textbf{0.8798} & \textbf{0.8827} & \textbf{0.7714} & \textbf{0.7820} & \textbf{0.7879} & \textbf{0.7916} & \textbf{0.6716} & \textbf{0.7030} & \textbf{0.7122} & \textbf{0.7244} \\
		\bottomrule[1pt]
	\end{tabular}
	
	\label{Tab:03}
\end{table*}

\textbf{mAP:} The results are presented in Table \ref{Tab:02}, which show that DMMVH is overall better than all the compared multi-view hashing methods by a large margin. For example, compared with the current state-of-the-art multi-view hashing method FGCMH, the average mAP score of our approach has increased by $3.51\%$, $9.58\%$, and $13.85\%$ on MIR-Flickr25K, NUS-WIDE, and MS COCO, respectively. That is, deep metric learning can indeed enhance the discriminative capability of hash codes. 

\textbf{Hash Code Length:} Intuitively, a longer hash code should preserve more semantic information and achieve better precision. Further, we study the effect of hash code length on multimedia retrieval mAP. The hash code is learned by setting the same code length for different methods. From Table \ref{Tab:02}, we notice that the mAP of our method increases as the hash code length grows. On the MS COCO dataset, our method obtains a performance improvement of $5.25\%$ when ranging hash code length from $16$ bits to $128$ bits. The experiments on other datasets show the same conclusion. However, some previous methods show a precision degradation while adding more hash bits, which indicates that these methods cannot scale well to hashing tasks with a longer length of hash code. On the contrary, our results demonstrate that the proposed method has a noticeable improvement in mAP as the length increases. Eventually, the experiments on the hyper-parameters are detailed in the appendix.

\subsection{Ablation Study}
\textbf{Experiment Settings:} To evaluate the effectiveness of our method, we perform an ablation study with different settings and report the performance. 
\begin{itemize}
	\item \emph{DMMVH-metric}: The quantization loss is removed. 
	\item \emph{DMMVH-quant}: The metric loss is removed. 
	\item \emph{DMMVH-image}: Only the visual features are used.
	\item \emph{DMMVH-text}: Only the text features are used.
	\item \emph{DMMVH-concat}: Image and text features are fused with concatenation.
	\item \emph{DMMVH}: Our full framework.
\end{itemize}

\textbf{Ablation Analysis:} The comparison results are presented in Table \ref{Tab:03}. Starting with the loss function, the quantization loss can not perform any optimization on the embeddings. The method retrieves data randomly, leading to terrible mAP across all the tasks. Deep metric loss, on the contrary, can help the method learn the embedding well. We notice that DMMVH-metric is slightly worse than the full method due to the lack of binarization constraint. From the view aspect, DMMVH-text is barely better than DMMVH-quant. DMMVH-image outperforms DMMVH-text in all tasks by a large margin indicating the image features contain more information than text. With concatenated multi-view features, our method already outperforms the state-of-the-art methods. But Context Gating further improves mAP. In addition, the comparison experiment of the old backbone network is detailed in the appendix.

\subsection{Convergence Analysis}
\begin{figure}
	\centering
	\subfigure{\includegraphics[scale=0.24]{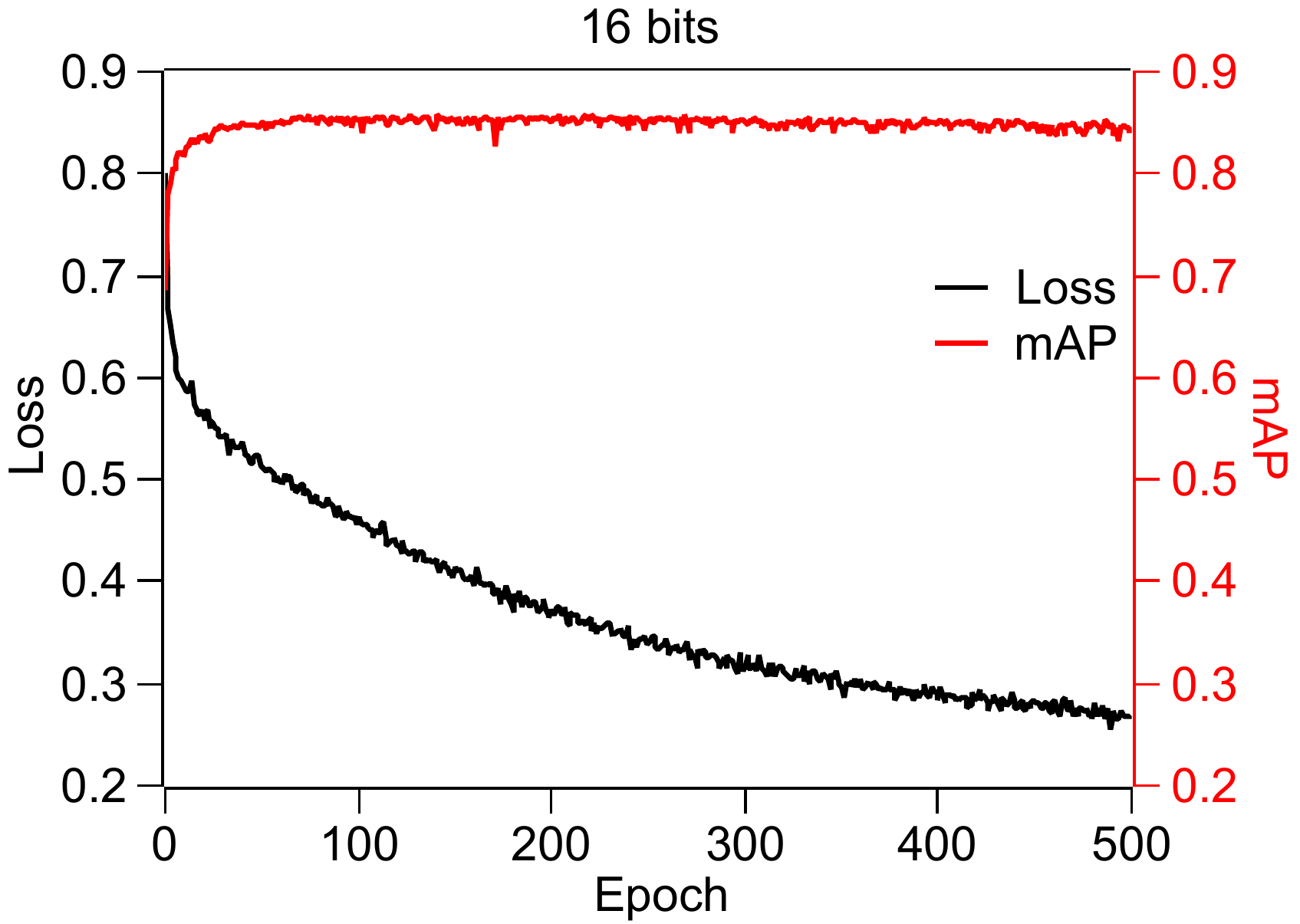}}
	\subfigure{\includegraphics[scale=0.24]{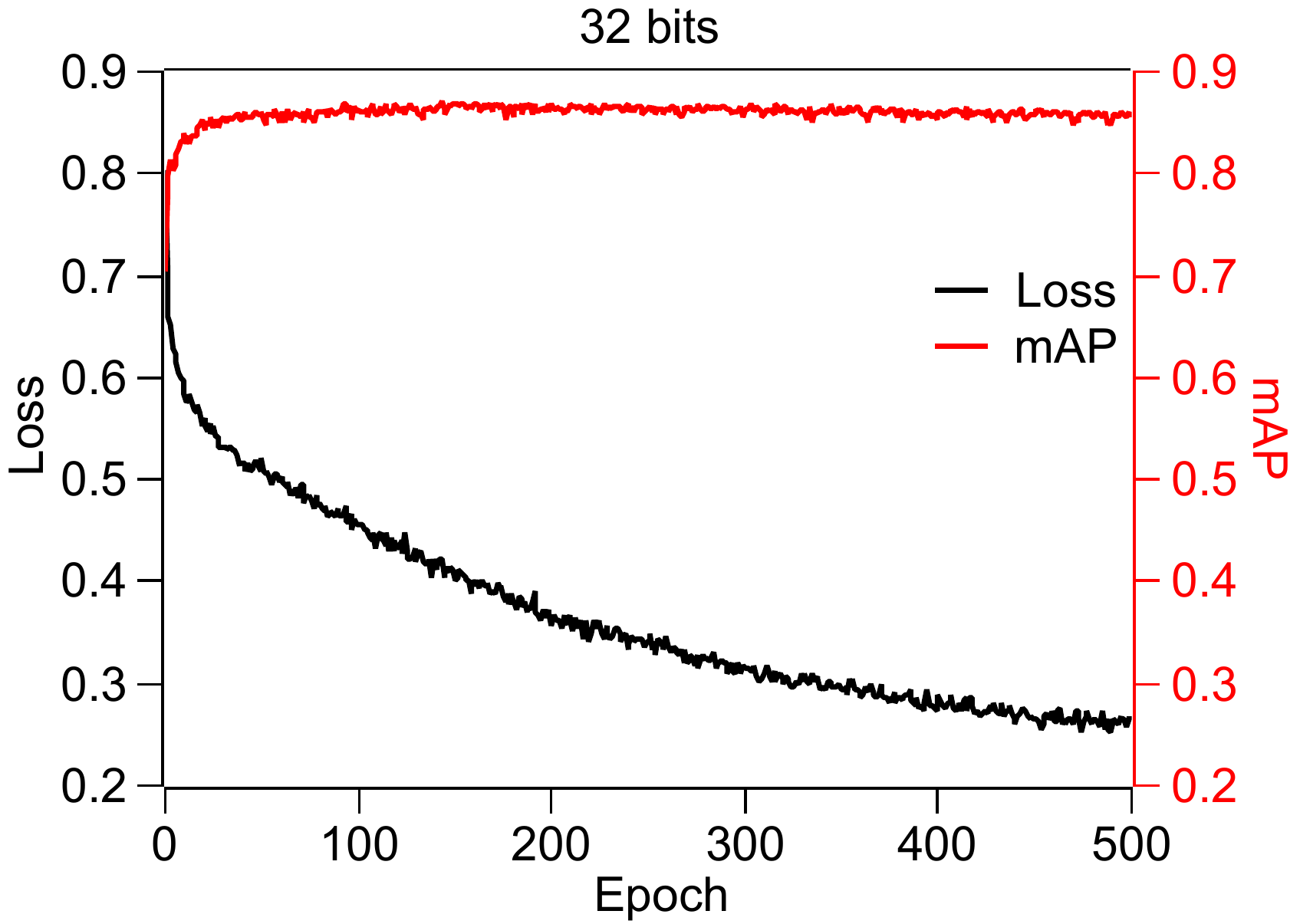}}
	\subfigure{\includegraphics[scale=0.24]{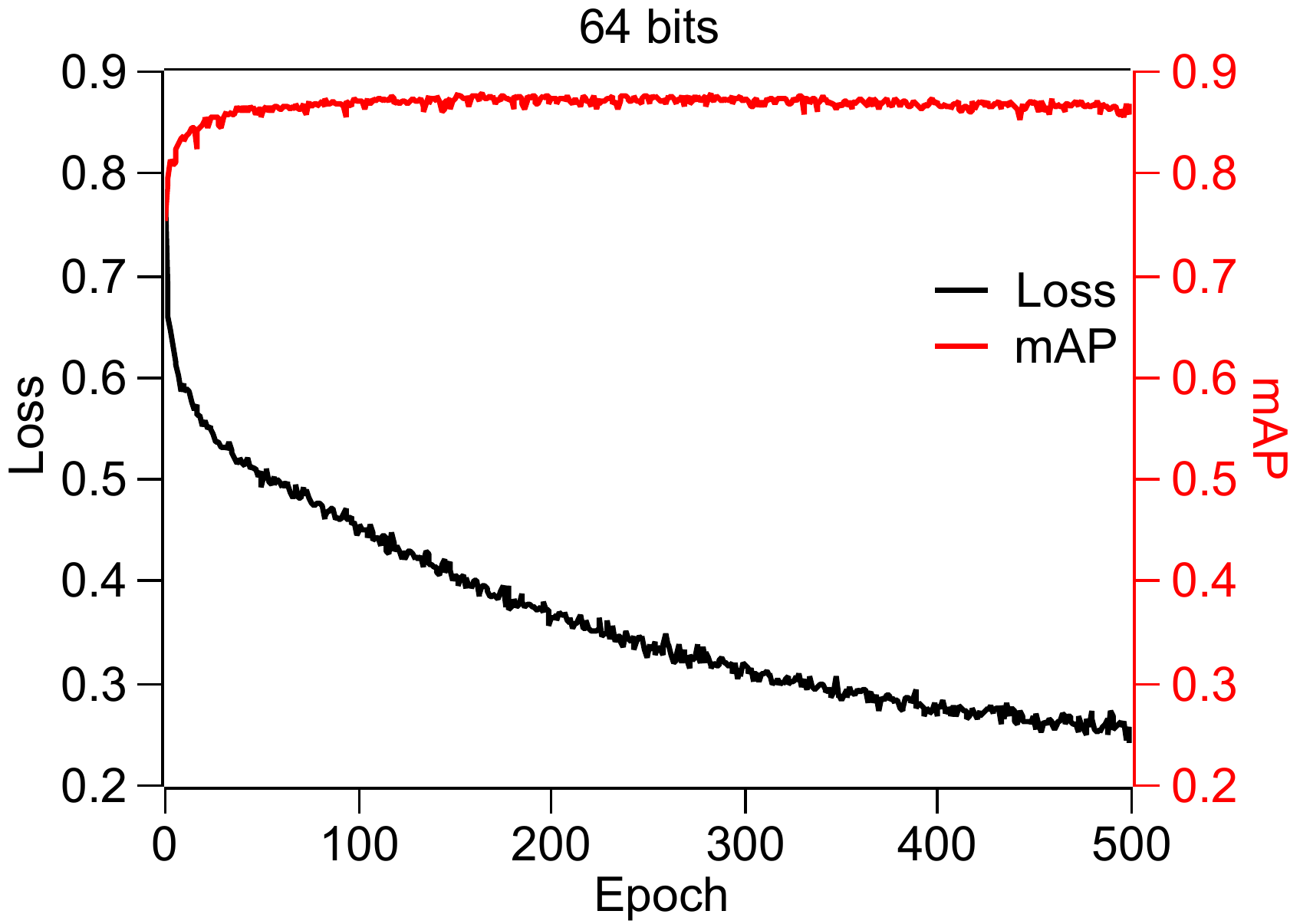}}
	\subfigure{\includegraphics[scale=0.24]{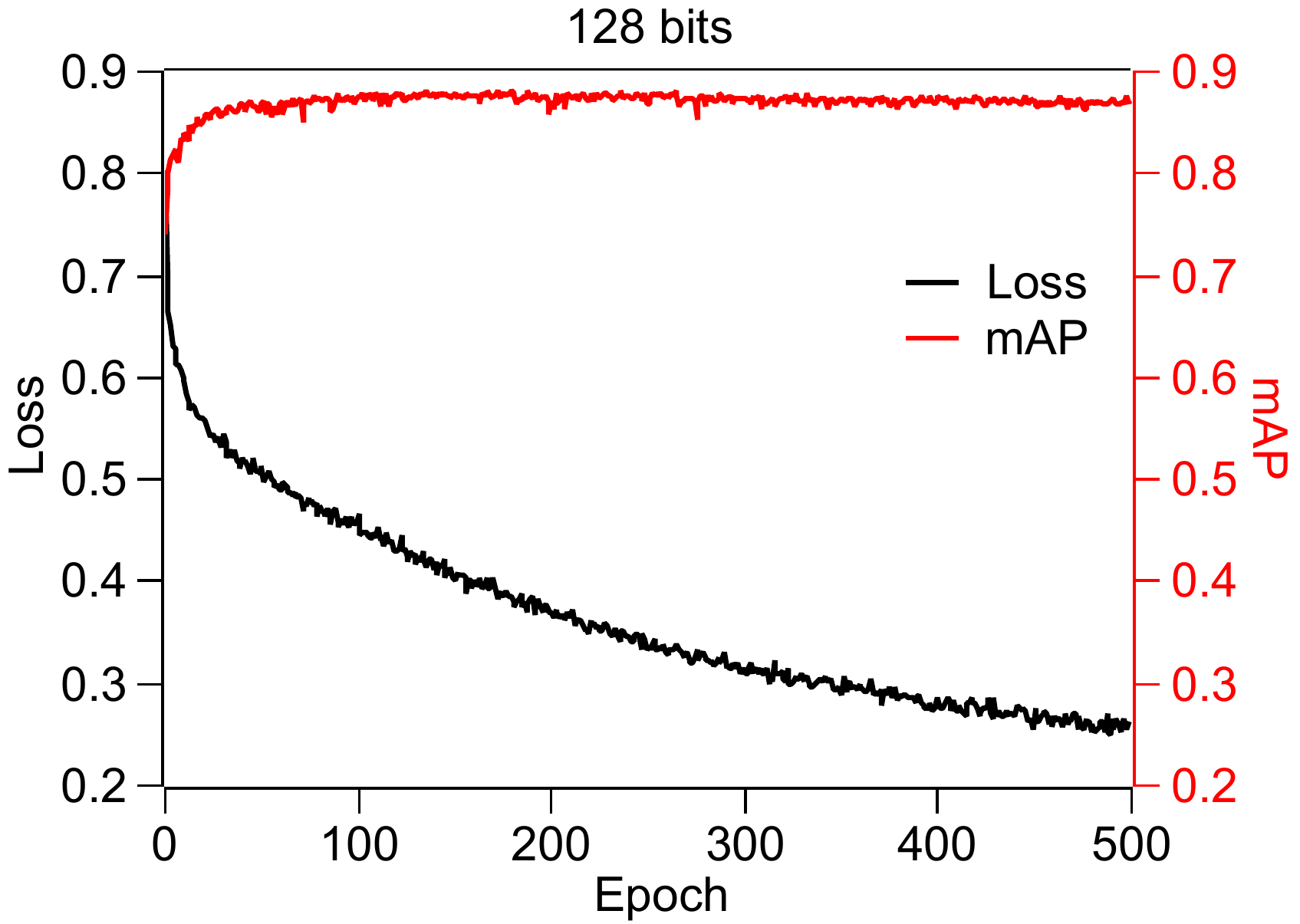}}
	\caption{The upper curve is the test mAP, and the bottom is the training loss on the MIR-Flickr25K dataset. }
	\label{fig:03}
\end{figure}

We conduct experiments to validate the convergence and generalization capability of DMMVH. We run hash benchmarks on the MIR-Flickr25K dataset in different code lengths. The results are shown in Fig. \ref{fig:03}. The figure delivers training loss and test mAP for analysis. As the training goes on, the loss gradually decreases. After 500 epochs, the loss becomes stable, which implies a local minimum is reached. For the test performance, the mAP goes up rapidly at the beginning of training. After 100 epochs, the test mAP stays stable. With further training, no degradation is observed on the test mAP, which indicates a good generalization capability. Similar convergence results are observed on other datasets.

\begin{figure}
	\centering
	\subfigure{\includegraphics[scale=0.24]{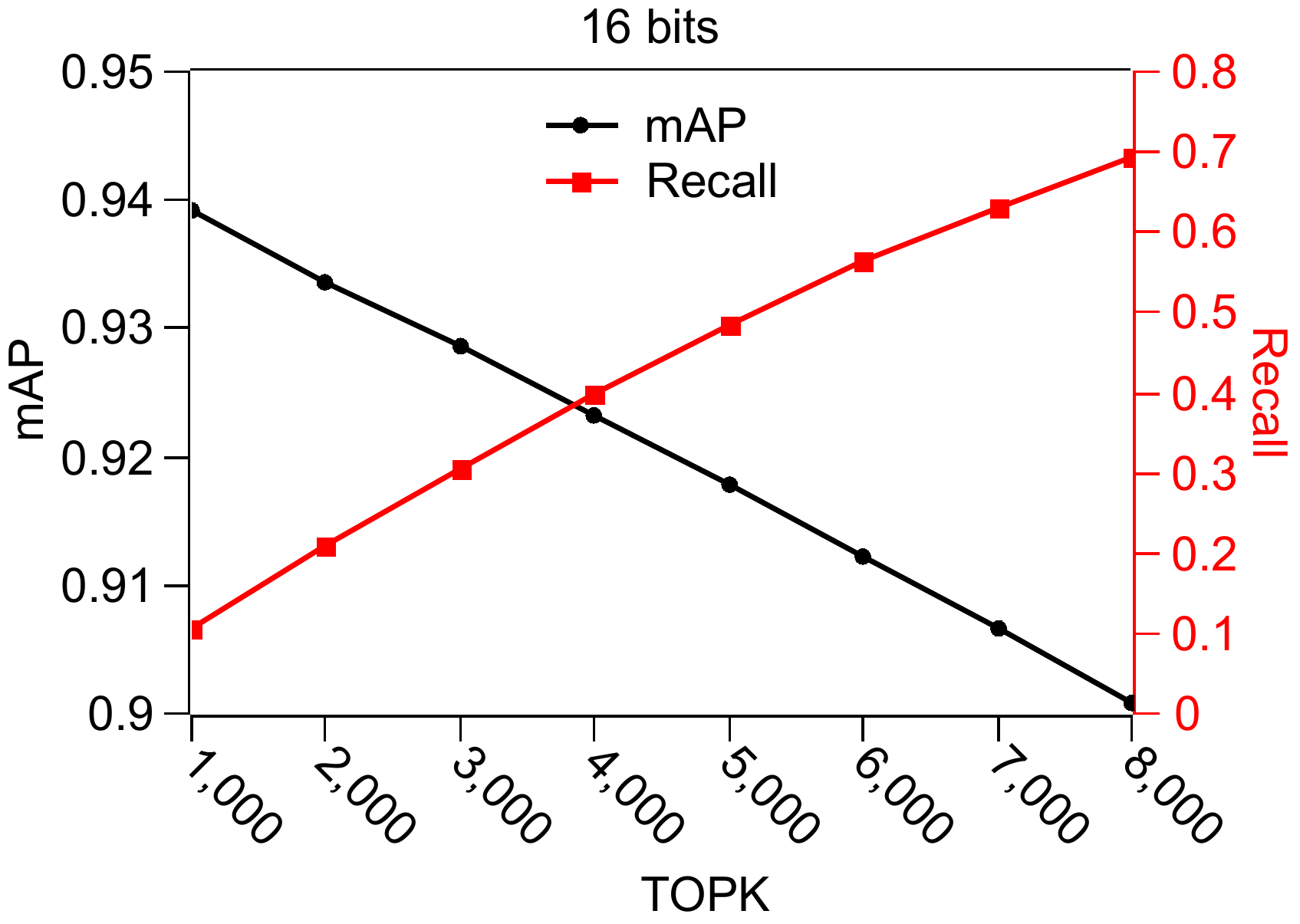}}
	\subfigure{\includegraphics[scale=0.24]{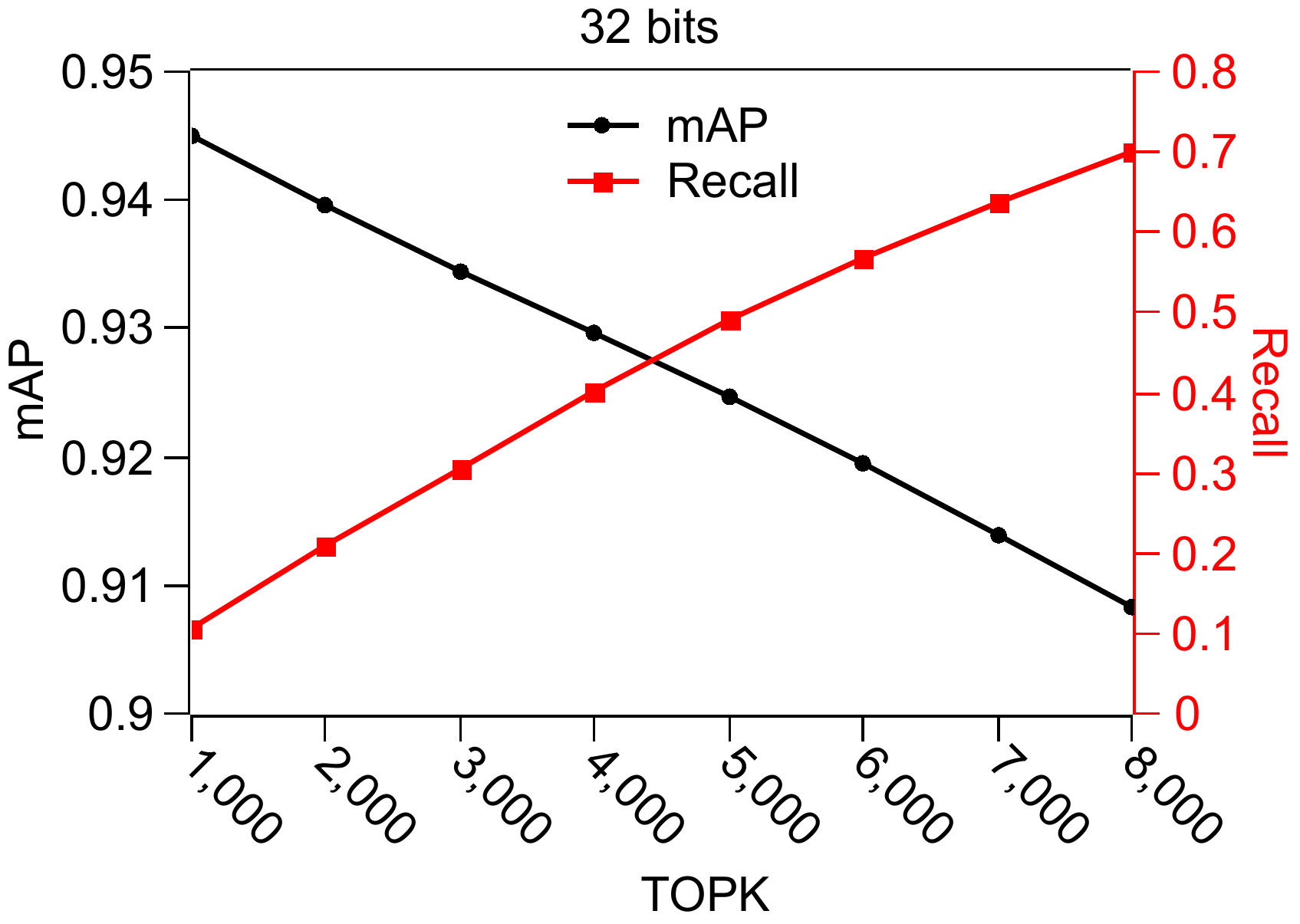}}
	\subfigure{\includegraphics[scale=0.24]{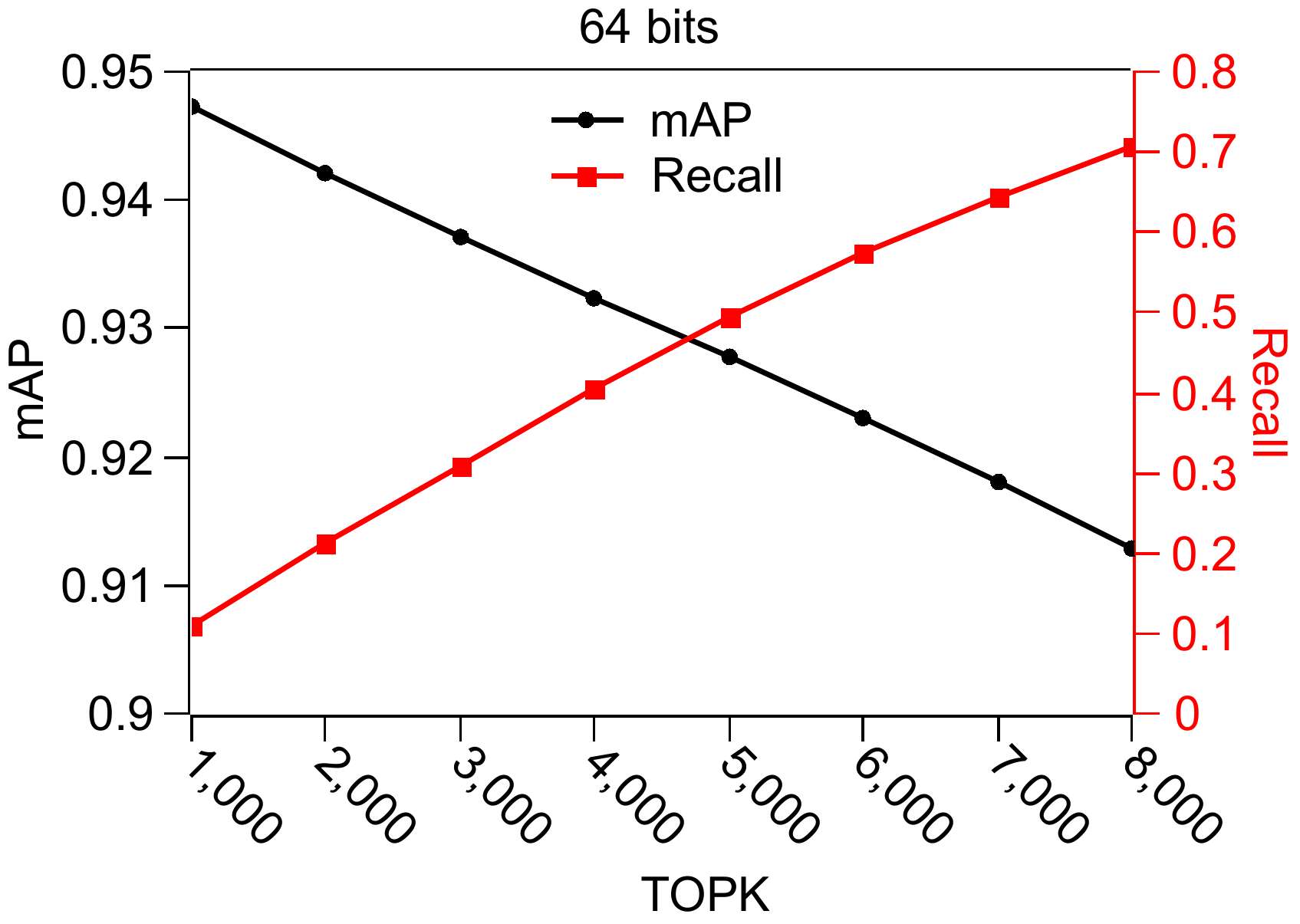}}
	\subfigure{\includegraphics[scale=0.24]{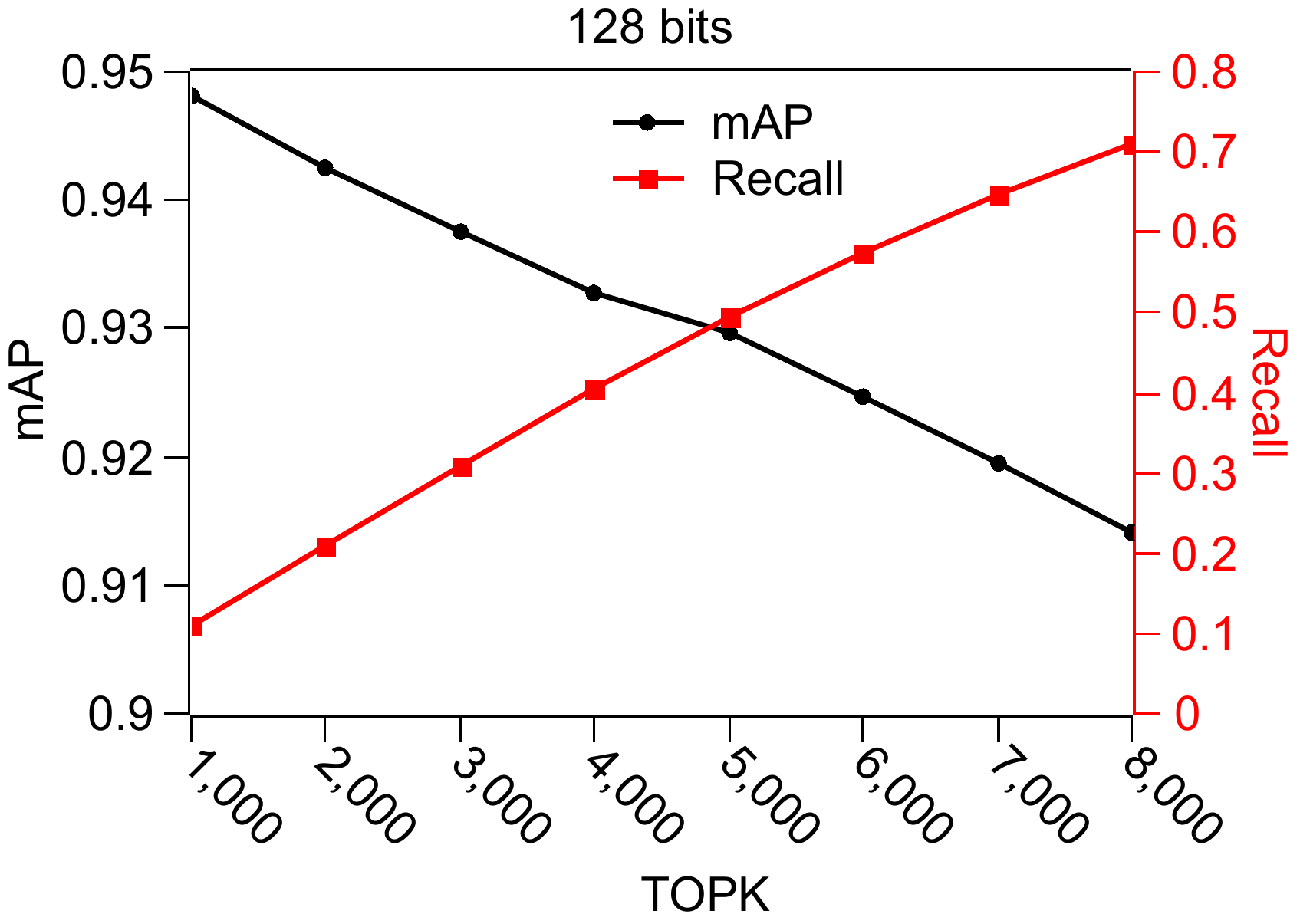}}
	\caption{The mAP@K and Recall@K curves on the MIR-Flickr25K dataset.}
	\label{fig:04}
\end{figure}

\subsection{mAP@K and Recall@K}
Fig. \ref{fig:04} shows the mAP@K and Recall@K curves with the increasing number of retrieval results on the MIR-Flickr25K dataset in different code lengths. The mAP of the four cases slightly decreases as $K$ increases, while the recall curve shows rapid linear growth. The tendency suggests that our method performs well in the retrieval tasks. Typical users only pay attention to a few results at the beginning of the retrieval results. Our method has even higher precision in this scenario. Experts tend to go through more results than typical users. Our approach can provide a linear growth recall as the number of retrieval results grows. Experts can expect consistent, high-quality results during their searches. To recap, DMMVH can deliver satisfying retrieval results for different user groups.

\section{Conclusion and Future Work}
We propose a new multi-view hashing framework (DMMVH). It introduces deep metric learning to solve multi-view hashing problems. We showed that DMMVH provides satisfying retrieval results to different types of users. Compared to typical graph-based methods, DMMVH is less computationally intensive. It utilizes Context Gating for multi-view features fusion and deep metric learning for representation optimization. The proposed method conquers two main challenges of the multi-view hashing problem. Under multiple experiment settings, it delivers up to $15.28\%$ performance gain over the state-of-the-art methods. In the experiment, we noticed some issues. For example, the performance gain is not quite significant as the length of the hash code increases. We will work on these issues to improve the proposed method further.

\section*{Acknowledgment}
This work is supported in part by the Zhejiang provincial ``Ten Thousand Talents Program'' (2021R52007), the National Key R\&D Program of China (2022YFB4500405), and the Science and Technology Innovation 2030-Major Project (2021ZD0114300).

\clearpage

\bibliographystyle{IEEEbib}
\bibliography{icme2023template}

\end{document}